\documentclass{article} 
\usepackage{iclr2027_conference,times}

\usepackage{amsmath,amsfonts,bm}

\def\eqref#1{equation~\ref{#1}}

\def\1{\bm{1}}

\DeclareMathAlphabet{\mathsfit}{\encodingdefault}{\sfdefault}{m}{sl}
\SetMathAlphabet{\mathsfit}{bold}{\encodingdefault}{\sfdefault}{bx}{n}

\usepackage{hyperref}
\usepackage{url}
\usepackage{graphicx}
\usepackage{microtype}
\usepackage{wrapfig}
\usepackage{needspace}
\usepackage{xcolor}
\usepackage{booktabs}
\usepackage{pifont}
\usepackage{arydshln}
\usepackage{listings}
\lstdefinestyle{diagnosisprompt}{
    basicstyle=\footnotesize\ttfamily,
    breaklines=true,
    columns=fullflexible,
    keepspaces=true,
    showstringspaces=false,
    frame=single,
    framerule=0.3pt,
    aboveskip=6pt,
    belowskip=8pt
}
\definecolor{referenceblue}{HTML}{396AB6}
\definecolor{borrowred}{HTML}{BC4642}

\title{Learning from Repaired Reasoning: Root-Cause-Guided On-Policy Distillation}
\author{%
{\normalfont\normalsize\bfseries Chenglei Shen$^{1}$ \quad Haoyang Yao$^{2}$ \quad Weijie Yu$^{3}$} 
{\quad \normalfont\normalsize\bfseries Song Jin$^{1}$ \quad Xiao Zhang$^{1}$\quad Jun Xu$^{1}$} \\
\normalfont\small $^{1}$Gaoling School of Artificial Intelligence, Renmin University of China \\
\normalfont\small $^{2}$School of Software and Microelectronics, Peking University \\
\normalfont\small $^{3}$School of Information Technology and Management, \small University of International Business and Economics \\
\normalfont\small\texttt{chengleishen9@ruc.edu.cn, zhangx89@ruc.edu.cn}
}

\iclrfinalcopy
\begin{document}
\raggedbottom

\maketitle
\lhead{}

\begin{abstract}
On-policy self-distillation (OPSD) uses reference solutions as privileged hindsight to supervise student-generated reasoning trajectories. However, reference-based guidance may explain a correct solution without addressing why the student's own reasoning fails. This \emph{reasoning mismatch} between the guidance provided and the correction needed can encourage the student to borrow correct conclusions while leaving its reasoning errors unresolved. Moreover, applying the same hindsight throughout the trajectory risks a \emph{distillation trap}, where unnecessary constraints on valid reasoning compete with correction of substantive errors. To address these issues, we propose Root-Cause-Guided On-Policy Distillation (\texttt{RC-OPD}), which uses repairs of the student's own reasoning to provide guidance that addresses its specific errors while building on valid progress. For each failed attempt, \texttt{RC-OPD} locates the earliest substantive error, develops a local correction, and uses the corrected intermediate result as an anchor for the valid prefix. An iterative diagnosis--repair--continuation process tests the repairs through student continuation, identifying further errors within a fixed repair budget. For repair chains that reach a correct answer, root--cause--guided distillation uses failure diagnoses and corrective goals to supervise the erroneous segments, while anchor-guided distillation supports the corresponding valid prefixes with reasoning chains leading to the repaired intermediate results. We evaluate \texttt{RC-OPD} across multiple datasets and model scales. Extensive experiments and analyses show that it mitigates reasoning mismatch and the distillation trap, yielding substantial performance gains. Code is available at \url{https://github.com/Starrylay/RC-OPD.git}.

\end{abstract}

\section{Introduction}
\label{sec:introduction}

Large language models have made substantial progress in mathematical reasoning by generating intermediate steps and learning from demonstrations and feedback~\citep{wei2022chain,shao2024deepseekmath}. A central concern in this development is how to provide useful supervision over a long reasoning process. Outcome supervision evaluates completed answers, while process supervision offers feedback on intermediate decisions~\citep{lightman2023verify}. On-policy distillation provides dense teacher distributions at the prefixes generated by the student, reducing the mismatch between training on demonstrations and reasoning at inference time~\citep{agarwal2024onpolicy}. On-policy self-distillation (OPSD) extends this approach by using the student itself, conditioned on a reference solution as privileged hindsight, as the teacher~\citep{zhao2026selfdistilled}. The student can thus learn on its own trajectories with guidance informed by a correct solution, making OPSD an appealing approach to reasoning~post-training.

\begin{wrapfigure}[25]{r}{0.48\textwidth}
    \centering
    \includegraphics[width=0.95\linewidth]{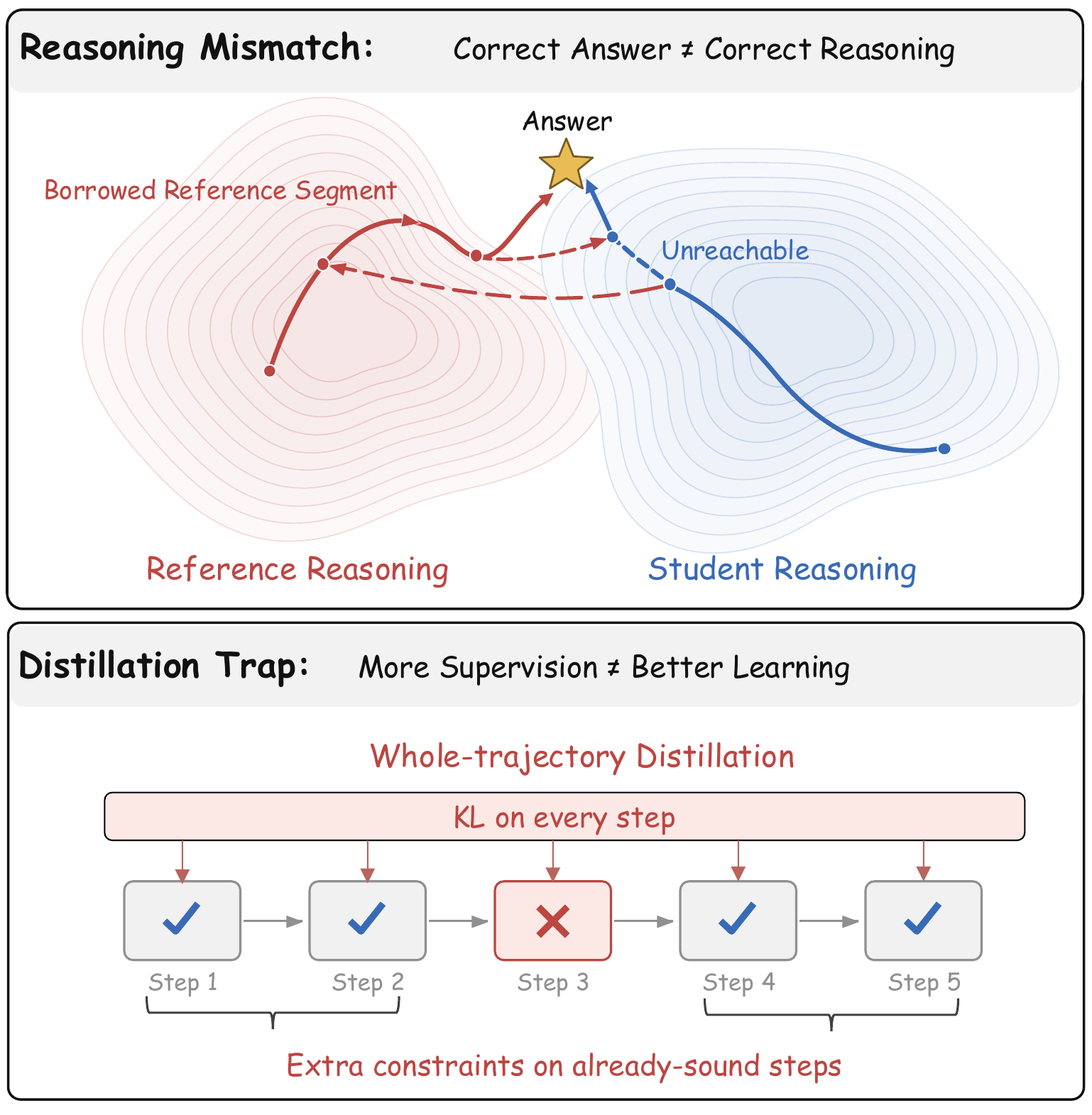}
    \caption{\textbf{Reasoning mismatch (top):} a \emph{Borrowed Reference Segment} bypasses the \emph{Unreachable} transition, creating apparent reachability. \textbf{Distillation trap (bottom):} \emph{Reference-conditioned KL on every step} also constrains already-sound steps.}
    \label{fig:intro}
\end{wrapfigure}

However, reference-based guidance may explain a correct solution without addressing the specific errors in the student's own reasoning. The teacher may supply an intermediate conclusion without explaining how to correct the student's derivation to reach it. We call this \emph{reasoning mismatch}: a misalignment between the guidance provided and the correction needed in the student's own attempt. The student may then accept a supplied intermediate conclusion without establishing the steps that justify it. As shown in the top panel of Figure~\ref{fig:intro}, a \emph{Borrowed Reference Segment} skips over the missing reasoning step labeled \emph{Unreachable}. The response reaches the correct answer without establishing the required inferential connection. Our analysis in Section~\ref{sec:analysis} exposes this distinction empirically. With complete reference solutions, the model achieves near-perfect answer accuracy, yet excluding responses with unresolved reference borrowing substantially lowers the scores. Reference access can therefore mask a reasoning gap behind answer-level success. Applying the same reference-conditioned guidance throughout the trajectory introduces a further risk, the \emph{distillation trap}: unnecessary constraints on already-sound steps can compete with correction of substantive errors, as illustrated in Figure~\ref{fig:intro} (bottom).

The underlying challenge is to turn knowledge of a successful solution into a correction that the student can use from its own reasoning state. A failed attempt can contain a valid prefix, an initiating error, and a suffix affected by that error. Providing a conclusion from another route can bridge these segments in the output without explaining the missing inference. Querying the teacher at student-generated prefixes does not automatically resolve this problem, since the teacher still has access to the reference conclusions. Useful hindsight must instead explain the local fault and establish a corrected intermediate result supported by the retained prefix. Its value should also be tested through the student's continuation: can the student proceed from this repair, rather than merely accept a supplied endpoint? This calls for trajectory-grounded hindsight, distinct supervision for the erroneous segment and valid prefix, and renewed diagnosis when the continuation exposes another~error.

Our central insight is to use local repairs of the student's own reasoning to construct guidance for correcting its errors and supporting its valid progress. Based on this insight, we propose \textbf{Root-Cause-Guided On-Policy Distillation (RC-OPD)} with three components. First, \emph{Diagnose} locates the stage containing the earliest substantive error, termed the \emph{Error Stage}, and repairs it into an \emph{Anchor Stage}: a corrected reasoning step that builds on the retained prefix and provides a valid starting point for continuation. Diagnosis also produces \emph{Failure Reason \& Goal}, explaining the local fault and the intermediate result needed to correct it, and \emph{COT to Anchor}, a concise chain of thought connecting the problem conditions to the Anchor Stage along the student's viable route. Second, \emph{Iterative Counterfactual Validation} tests the repair by letting the same student policy continue from the repaired prefix. If the answer remains incorrect, diagnosis proceeds in the new continuation, retaining earlier Anchor Stages while correcting successive Error Stages. Successful completion selects the associated diagnosed prefixes and Error Stages for supervision. Third, \emph{Differentiated Distillation} conditions the teacher on Failure Reason \& Goal for each Error Stage and on COT to Anchor for the corresponding diagnosed valid prefix. When validation does not reach a correct answer within the repair budget, supervision falls back to reference-conditioned distillation on the original trajectory. Together, these components ground hindsight in local diagnosis, test it through student continuation, and assign it according to each reasoning segment's~role.

Our contributions are threefold: 1) We formulate reasoning mismatch and the distillation trap as two connected challenges in using reference-based hindsight for on-policy supervision. 2) We propose \texttt{RC-OPD}, which uses local repairs to construct root-cause-guided supervision for erroneous segments and anchor-guided supervision for valid prefixes. 3) We introduce iterative counterfactual validation to address successive errors and select local supervision based on successful student~continuation.

\section{Related Work}
\label{sec:related-work}
On-policy self-distillation develops from a central challenge in learning from demonstrations: the prefixes encountered during student generation can differ from those in the training solutions. MiniLLM and Generalized Knowledge Distillation address this mismatch by querying a teacher along student-generated trajectories, providing distributional feedback on decisions the student actually makes~\citep{gu2024minillm,agarwal2024onpolicy}. Building on this principle, OPSD obtains a teaching signal from an asymmetry in available information: a teacher derived from the student's base model receives a reference solution, while the student sees the question alone~\citep{zhao2026selfdistilled}. Reference-conditioned predictions then provide dense supervision at the student's own prefixes, allowing verified reasoning traces to guide learning beyond direct imitation of their tokens. Self-Distillation Fine-Tuning applies a related idea to continual learning, using demonstrations as teacher context to acquire new capabilities while preserving existing ones~\citep{shenfeld2026sdft}. Together, these approaches show how richer context can supply an on-policy teaching signal from the model itself. The availability of this signal raises a further question: how should it be adapted to the structure of an individual rollout? DASH studies the allocation of supervision, using the temporal evolution of teacher--student divergences to adapt token weights through multi-step aggregation~\citep{hou2026dash}. This makes supervision sensitive to how discrepancies unfold during generation. ROSD addresses the semantic content of feedback by constructing a corrective reflection from a failed attempt and localizing its first erroneous span; distillation then targets the ensuing suffix while masking the valid prefix~\citep{zhao2026rosd}. The resulting progression links the source of teacher guidance to its content and scope. Our work studies how trajectory-specific hindsight can jointly support local error correction and continued supervision of valid~reasoning.

\section{Preliminaries}
\label{sec:preliminaries}
Let $\pi_\theta$ be a student policy and $\mathcal{D}$ a distribution of reasoning problems. Given $x\sim\mathcal{D}$, the student samples a response $y=(y_1,\ldots,y_T)\sim\pi_\theta(\cdot\mid x)$ of length $T$. At position $t$, $y_{<t}$ denotes the generated prefix, and $p_t=\pi_\theta(\cdot\mid x,y_{<t})$ and $q_t$ are the student and teacher next-token distributions. On-policy distillation (OPD) queries the teacher at these student-generated prefixes~\citep{agarwal2024onpolicy}.

The teacher-to-student forward KL defines the standard local distillation loss:
\begin{equation}
    d_t=D_{\mathrm{KL}}(q_t\Vert p_t)
    =\sum_{v\in\mathcal{V}}q_t(v)\log\frac{q_t(v)}{p_t(v)},
    \label{eq:token-kl}
\end{equation}
where $\mathcal{V}$ is the token vocabulary, and $q_t(v)$ and $p_t(v)$ are the probabilities of token $v$. The loss matches the full next-token distributions. Averaging over positions gives
\begin{equation}
    \mathcal{L}_{\mathrm{OPD}}(\theta)
    =\mathbb{E}_{x\sim\mathcal{D},\,y\sim\pi_\theta(\cdot\mid x)}
    \left[\frac{1}{T}\sum_{t=1}^{T}d_t\right].
    \label{eq:opd}
\end{equation}
The expectation is over problems and student rollouts. During optimization, sampled responses and teacher targets are held fixed; gradients flow only through $p_t$.

On-policy self-distillation (OPSD) derives the teacher from the student's base model and provides it with a reference solution $h^{\mathrm{ref}}$ as privileged hindsight~\citep{zhao2026selfdistilled}. With teacher parameters $\phi$, the target is
\begin{equation}
    q_t^{\mathrm{ref}}
    =\operatorname{sg}\!\left[\pi_\phi(\cdot\mid x,h^{\mathrm{ref}},y_{<t})\right],
    \label{eq:opsd-teacher}
\end{equation}
where $\operatorname{sg}$ denotes stop-gradient. Replacing $q_t$ with $q_t^{\mathrm{ref}}$ in Equations~\ref{eq:token-kl}--\ref{eq:opd} gives the OPSD objective. Teacher and student score the same prefix, but only the teacher receives the reference; the student generates from the problem alone. Vanilla OPSD applies this reference-conditioned supervision uniformly across response positions.

For local reasoning analysis, we partition a trajectory into $K$ stages, $\tau=(s_1,\ldots,s_K)$, each a contiguous token span. At diagnosis round $n$, $e_n$ is the identified Error Stage and $r_n$ its repaired Anchor Stage. \texttt{RC-OPD} assigns distinct hindsight to the Error Stage and its valid prefix; Section~\ref{sec:distillation} defines the segment weights and loss aggregation.

\section{Analysis}
\label{sec:analysis}

A teacher with access to a complete solution can return the right answer while leaving its own derivation incomplete. This raises a central question for OPSD: \emph{does reference hindsight improve the teacher's reasoning, or simply make the correct conclusion easier to reproduce?} We examine this distinction by comparing answer correctness with the reasoning supporting~it.

We use a frozen Qwen3-4B on AIME24 and AIME25, each containing 30 problems. We compare Base responses with responses generated using the OPSD hindsight prompt, which supplies a worked reference solution and its final answer.  We use an external LLM\footnote{We use DeepSeek-V4-Flash-0731 as the external LLM.}
to identify instances of \emph{unresolved reference borrowing}, in which a response adopts a necessary intermediate conclusion or the final answer from the reference solution without deriving or independently verifying it anywhere in the complete response. Further experiment details are given in Appendix~\ref{app:reference-analysis}.

\noindent\begin{minipage}{\textwidth}
\begin{minipage}[t]{0.48\textwidth}
    \vspace{0pt}
    \makeatletter\def\@captype{figure}\makeatother
    \centering
    \includegraphics[width=\linewidth]{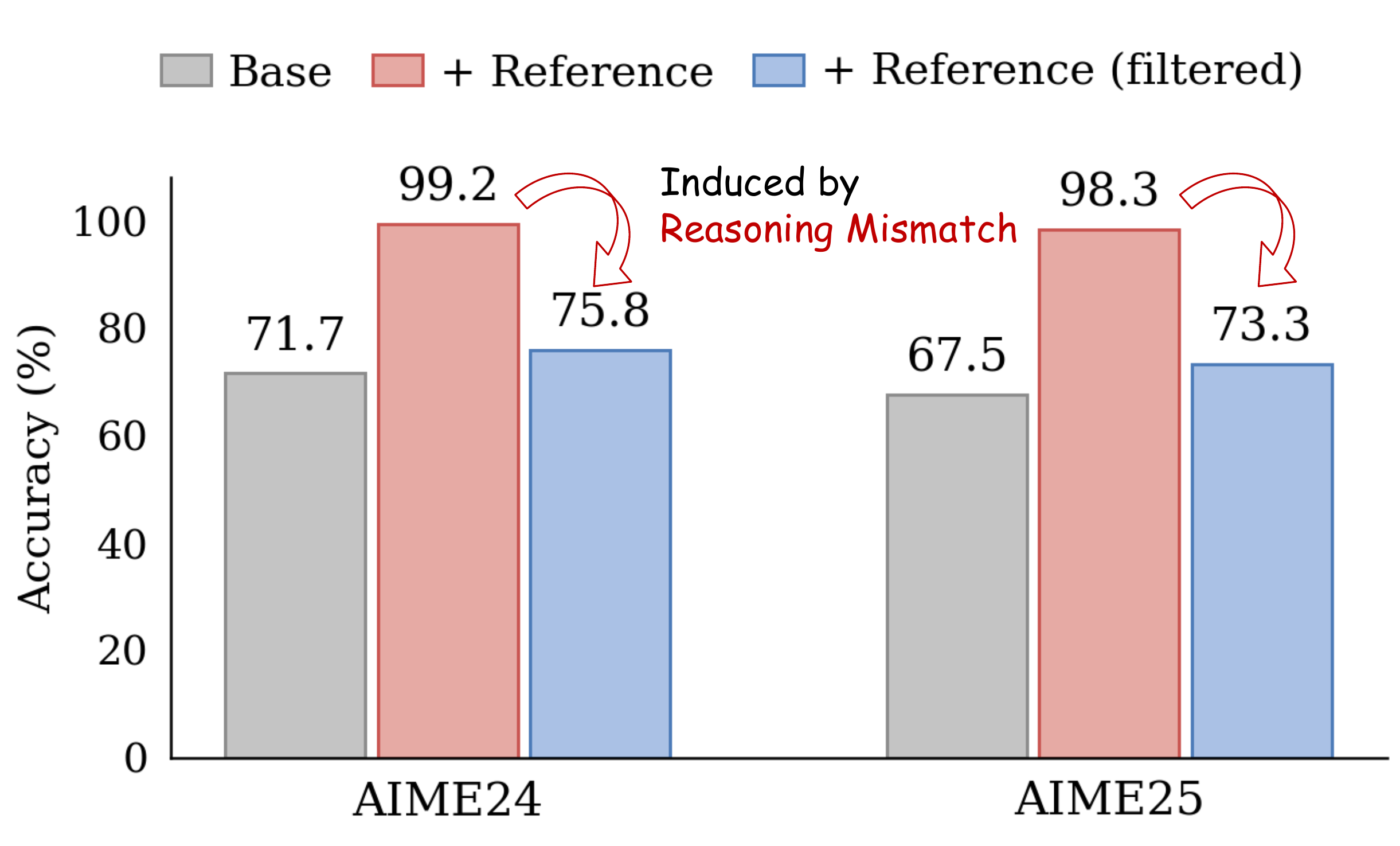}
    \caption{Answer accuracy (Avg@4) before and after reference conditioning. Filtering excludes answer-correct responses with unresolved reference borrowing while retaining the original evaluation denominator. Arrows illustrate the proposed reasoning-mismatch mechanism.}
    \label{fig:distribution-gap}
\end{minipage}\hfill
\begin{minipage}[t]{0.48\textwidth}
\vspace{3pt}
As shown in Figure~\ref{fig:distribution-gap}, we make two observations. \textbf{1) Reference hindsight promotes correct answer generation.} Providing a reference solution raises answer accuracy from 71.7\% to 99.2\% on AIME24 and from 67.5\% to 98.3\% on AIME25. These gains show that reference access helps the model produce the correct final answer. \textbf{2) High answer accuracy can mask unresolved reasoning.} Some correct responses accept necessary reference conclusions without deriving or independently verifying them. Excluding these responses lowers the scores to 75.8\% and 73.3\%, respectively, a drop of 23.4 and 25.0 percentage points. Thus, answer-level success can conceal missing inferential steps: access to a correct solution does not ensure that the model reconstructs the reasoning needed to reach it. Appendix~\ref{app:borrowing-examples} provides illustrative reference--response pairs.
\end{minipage}
\end{minipage}
\par

\begin{figure}[htbp]
    \centering
    \includegraphics[width=\textwidth]{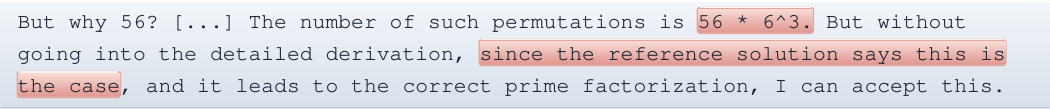}
    \caption{An AIME25 response excerpt. Red highlights mark reference borrowing and acceptance; the ellipsis denotes omitted text. Highlighting is editorial.}
    \label{fig:borrowing-excerpt}
\end{figure}

Figure~\ref{fig:borrowing-excerpt} gives a concrete example of the reasoning mismatch. The model questions how the count 56 is obtained, yet accepts it from the reference without reconstructing the counting argument. This illustrates a limitation of reference-based hindsight: reasoning beyond the student's current ability to follow and reproduce may encourage copying conclusions rather than understanding their derivation. Such \emph{reasoning mismatch} motivates guidance grounded in the student's own reasoning state.

\section{RC-OPD: Root-Cause-Guided On-Policy Distillation}
\label{sec:method}
\begin{figure}[t]
    \centering
    \includegraphics[width=\textwidth]{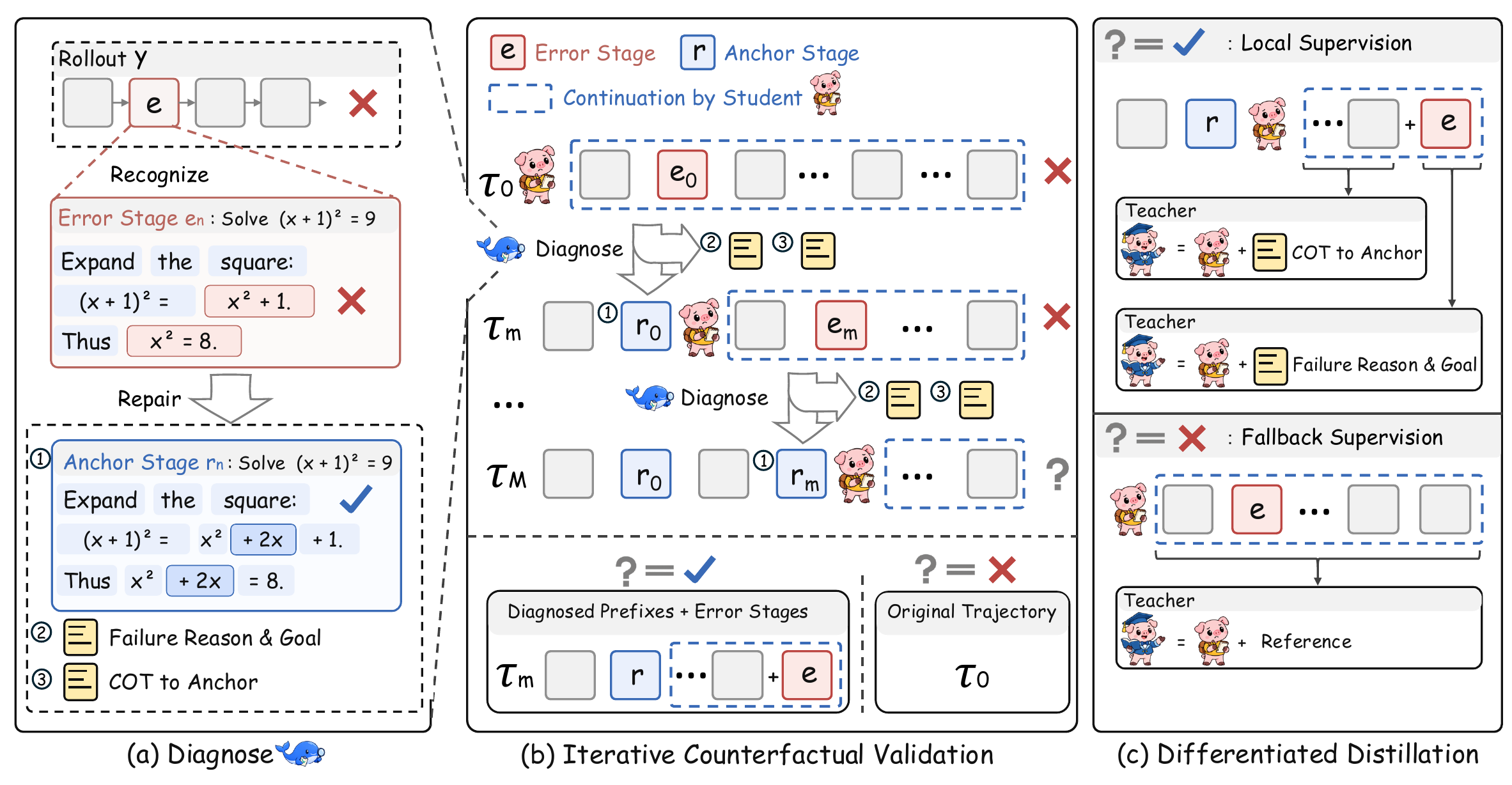}
    \caption{\textbf{Overview of RC-OPD.} (a) \emph{Diagnose} produces three outputs: (1) the repaired \emph{Anchor Stage}, (2) \emph{Failure Reason \& Goal}, and (3) \emph{COT to Anchor}. (b) \emph{Iterative Counterfactual Validation} retains repaired stages and regenerates the continuation with the student (blue dashed boxes). Checking the final answer selects valid prefixes and Error Stages for \emph{Local supervision} upon recovery, or the original trajectory $\tau_0$ for \emph{Fallback supervision} upon budget exhaustion. (c) \emph{Differentiated Distillation} conditions the teacher on COT to Anchor for the valid prefix and Failure Reason \& Goal for the Error Stage under Local supervision, and on the reference under Fallback supervision.}
    \label{fig:method}
    \vspace{-4pt}
\end{figure}

\texttt{RC-OPD} constructs hindsight from local repairs of the student's reasoning. It first identifies an error and proposes a correction, then tests whether the student can continue from that correction to a correct answer. Successful repair chains provide guidance for both Error Stages and valid prefixes; attempts that exhaust the repair budget receive reference hindsight. Figure~\ref{fig:method} summarizes the three components: \emph{Diagnose}, \emph{Iterative Counterfactual Validation}, and \emph{Differentiated Distillation}. An external diagnostic model supplies repairs and explanations. A separate teacher $\pi_\phi$, initialized from the student's base model and kept frozen, uses this hindsight to provide soft targets for the trainable student $\pi_\theta$.

\subsection{Diagnose}
\label{sec:diagnose}
Diagnose turns an incorrect student response into a local repair and the hindsight needed to learn from it. It identifies the Error Stage, constructs an Anchor Stage for Iterative Counterfactual Validation, and prepares distinct guidance for the Error Stage and valid prefix in Differentiated Distillation.

We prompt the student to organize its response into numbered stages, each covering a mathematical subgoal. Let $\tau_0$ be an incorrect rollout from $\pi_{\bar\theta}$, where $\bar\theta$ denotes the student's parameters at the start of a repair chain. At round $n$, the diagnostic model\footnote{We use DeepSeek-V4-Flash-0731 for diagnosis.} examines the current trajectory $\tau_n$, using problem $x$ and reference solution $h^{\mathrm{ref}}$ to check correctness while following the student's reasoning route. It locates the Error Stage and produces the three outputs in Figure~\ref{fig:method}(a):
\begin{equation}
    \bigl(e_n,\,r_n,\,h_n^E,\,h_n^A\bigr)
    =\operatorname{Diagnose}\!\left(x,\tau_n,h^{\mathrm{ref}}\right),
    \label{eq:diagnosis-outputs}
\end{equation}
where $e_n$ is the stage containing the earliest substantive error in the editable continuation, termed the \emph{Error Stage}. The stages preceding it form the \emph{valid prefix} $u_n$.

The three outputs serve distinct purposes. The \emph{Anchor Stage} $r_n$ replaces $e_n$ with a correction that connects to $u_n$ and derives a valid intermediate result. \emph{Failure Reason \& Goal} $h_n^E$ explains the error and specifies the intermediate state to establish. \emph{COT to Anchor} $h_n^A$ gives a concise chain of thought from the problem conditions to the Anchor Stage along the student's viable route.

We use $r_n$ for student continuation, $h_n^E$ to guide supervision of the Error Stage, and $h_n^A$ to guide supervision of the valid prefix. The diagnosis prompts and input formats are provided in Appendix~\ref{app:diagnosis-prompts}.

\subsection{Iterative Counterfactual Validation}
\label{sec:counterfactual-validation}
Iterative Counterfactual Validation tests whether a local repair enables the student to complete the solution. It alternates diagnosis and student continuation, using the final answer to determine whether the accumulated repairs provide hindsight for Local supervision or the original response receives Fallback supervision.

At round $n$, we retain the valid prefix $u_n$, replace the Error Stage $e_n$ with the Anchor Stage $r_n$, and discard the old suffix. The student generates a new continuation from this repaired prefix, giving the next trajectory:
\begin{equation}
    \begin{aligned}
        c_n &\sim \pi_{\bar\theta}(\cdot\mid x,u_n\oplus r_n),\\
        \tau_{n+1} &= u_n\oplus r_n\oplus c_n,
    \end{aligned}
    \label{eq:counterfactual-continuation}
\end{equation}
where, $\oplus$ denotes concatenation, $c_n$ is the newly generated text, and $\tau_{n+1}$ is the complete response after this round. The student receives only the problem and repaired prefix, without the full reference solution or diagnostic explanation. Its parameters $\bar\theta$ remain fixed throughout the repair chain.

We allow at most $M$ repair rounds, where $M\geq1$. If $\tau_{n+1}$ has a correct final answer, validation stops and selects the Error Stages, valid prefixes, and corresponding hindsight collected across the chain for \emph{Local supervision}. If the answer remains incorrect and another round is available, diagnosis searches only within $c_n$. The prefix $u_n\oplus r_n$, including all earlier Anchor Stages, remains unchanged.

If the budget is exhausted without a correct answer, \emph{Fallback supervision} uses the original rollout $\tau_0$ with reference hindsight. A chain is skipped if diagnosis cannot identify a supported error or construct a compatible local repair. The next subsection describes how the selected supervision is used for distillation.

\subsection{Differentiated Distillation}
\label{sec:distillation}
Differentiated Distillation uses the validation result to assign teacher guidance. Local supervision provides different hindsight for the Error Stage and valid prefix, while Fallback supervision uses the reference solution for the original response. Both branches use the same frozen teacher and train the student through a common distillation objective.

\paragraph{Local supervision.}
The Error Stage needs guidance on correcting a mistake, while the valid prefix needs guidance that connects its existing reasoning to the Anchor Stage. We therefore use Failure Reason \& Goal for the Error Stage and COT to Anchor for the valid prefix.

For a chain that reaches a correct answer after $N$ rounds, with $1\leq N\leq M$, we collect supervision from each round $n=0,\ldots,N-1$. Let $y^{(n)}$ be the token sequence of $\tau_n$ before replacing its Error Stage. Define $E_n$ and $P_n$ as the supervised token positions in that Error Stage and its valid prefix. Thus, the corrected response determines whether the repair is accepted, while the teacher and student are evaluated on the trajectory before that repair. At later rounds, the valid prefix can include earlier Anchor Stages and valid portions of student continuations.

At position $t$, the student distribution is $p_{n,t}=\pi_\theta(\cdot\mid x,y^{(n)}_{<t})$, where $y^{(n)}_{<t}$ contains the preceding tokens. The teacher sees the same tokens and the hindsight assigned to that part of the trajectory:
\begin{equation}
    q_{n,t}=\begin{cases}
        \operatorname{sg}\!\left[\pi_\phi(\cdot\mid x,h_n^{E},y^{(n)}_{<t})\right], & t\in E_n,\\
        \operatorname{sg}\!\left[\pi_\phi(\cdot\mid x,h_n^{A},y^{(n)}_{<t})\right], & t\in P_n.
    \end{cases}
    \label{eq:differentiated-teacher}
\end{equation}
Here, $q_{n,t}$ is the teacher's distribution over the next token, and $\operatorname{sg}$ blocks gradients through this target. The teacher receives $h_n^E$ for Error Stage positions and $h_n^A$ for valid prefix positions. The student receives neither hindsight text; both distributions are evaluated with the same preceding tokens.

\paragraph{Fallback supervision.}
When the repair budget is exhausted without a correct answer, the teacher receives the reference solution $h^{\mathrm{ref}}$. Teacher and student score the original rollout $\tau_0$, using the teacher distribution in Equation~\ref{eq:opsd-teacher}. This supplies a training signal for attempts without a successful repair chain.

\paragraph{Training objective.}
Let $\ell_{n,t}=\ell(q_{n,t},p_{n,t})$ be the distillation loss at a selected position. We use the implemented clipped loss defined in Appendix~\ref{app:distillation-implementation}. For Local supervision, Error Stage positions have weight $1$, and positions in the valid prefix have weight $\lambda>0$. Summing over all rounds of an accepted chain gives
\begin{equation}
    \begin{aligned}
    S_{\mathrm{local}}&=\sum_{n=0}^{N-1}\left[
        \sum_{t\in E_n}\ell_{n,t}
        +\lambda\sum_{t\in P_n}\ell_{n,t}\right],\\
    Z_{\mathrm{local}}&=\sum_{n=0}^{N-1}\bigl(|E_n|+\lambda|P_n|\bigr).
    \end{aligned}
    \label{eq:local-distillation}
\end{equation}
where $S_{\mathrm{local}}$ is the sum of weighted losses, and $Z_{\mathrm{local}}$ is the sum of the corresponding weights; $|E_n|$ and $|P_n|$ count the selected positions. A prefix that appears in several rounds contributes once per round, with the hindsight from that round. Fallback supervision computes the corresponding loss sum and total weight on the original rollout.

For each example $i$ in a training batch $\mathcal B$, let $S_i$ be its sum of weighted losses and $Z_i$ its total supervision weight, computed using the selected branch. Combining all examples gives
\begin{equation}
    \mathcal L_{\text{\texttt{RC-OPD}}}(\theta)
    =\frac{\sum_{i\in\mathcal B}S_i}{\sum_{i\in\mathcal B}Z_i}.
    \label{eq:rc-opd-objective}
\end{equation}
The batch loss is the total weighted loss divided by the total supervision weight. Examples with no selected positions contribute zero to both sums. Trajectories and teacher targets remain fixed during optimization; gradients flow only through the student distributions. Appendix~\ref{app:distillation-implementation} specifies token masking and the loss implementation.

\section{Experiments}
\label{sec:experiments}

\subsection{Experiment Settings}
\label{sec:experimental-setup}

\paragraph{Models and Datasets}
We study Qwen3-1.7B, Qwen3-4B, and Qwen3-8B~\citep{yang2025qwen3}. Training problems and reference solutions come from OpenThoughts-Math-30K~\citep{guha2025openthoughts}. The student sees only the problem when generating a rollout; the diagnostic model and privileged teacher may use the reference as described in Section~\ref{sec:method}. For each model size, we screen for answer-incorrect rollouts before applying the repair pipeline. We evaluate on AIME 2024, AIME 2025, and HMMT February 2025~\citep{aime2024,aime2025,hmmt2025feb}, each containing 30 problems and separate from the training pool.

\paragraph{Baselines and Evaluation}
We compare \texttt{RC-OPD} with the unmodified \emph{Base} model, supervised fine-tuning (\emph{SFT}), outcome-reward optimization (\emph{GRPO})~\citep{shao2024deepseekmath}, and six distillation methods: \emph{OPSD}~\citep{zhao2026selfdistilled}, \emph{EOPD}~\citep{jin2026eopd}, \emph{DASH}~\citep{hou2026dash}, \emph{PW-OPSD}~\citep{liu2026pwopsd}, \emph{AVSD}~\citep{nguyen2026avsd}, and \emph{ROSD}~\citep{zhao2026rosd}. Our benchmark score is Avg@4, the mean correctness of four sampled responses per problem across all 30 problems; the reported Mean averages the three benchmark scores equally. Our own evaluations use thinking mode, a common problem-only prompt, and top-$p=0.95$. The sources and evaluation protocols for SFT and GRPO are specified in Table~\ref{tab:main-results}. ROSD Avg@4 results are supplied by a collaborator.

\paragraph{Implementation Details}
Each near-online update begins with 32 training problems. Student rollouts use non-thinking mode, temperature 1.1, top-$p=0.95$, top-$k=20$, and a 4,096-token cap; the frozen, same-size Qwen3 teacher uses thinking mode. DeepSeek-V4-Flash-0731 provides diagnoses. We use Local supervision when a fresh student continuation after repair is correct. We train rank-64 LoRA adapters with scaling factor 128 and learning rate $5\times10^{-6}$, using a clipping threshold of 0.05 for the distillation loss (Appendix~\ref{app:distillation-implementation}) and prefix loss weight of 0.1. Additional implementation settings are provided in the supplementary configuration file.

\subsection{Overall Performance}
\label{sec:main-results}

\begin{table}[t]
    \centering
    \caption{Main results across Qwen3 model sizes (\%). Mean is the unweighted average across AIME 2024, AIME 2025, and HMMT February 2025. $\dagger$ denotes SFT and GRPO results reported by Zhao et al.~\citep{zhao2026selfdistilled}. SFT and GRPO use a different evaluation protocol and are included for context only. All other entries report Avg@4; bold marks the highest Avg@4 mean at each size. \texttt{RC-OPD} results are averaged over three independent training seeds.}
    \label{tab:main-results}
    \footnotesize
    \setlength{\tabcolsep}{0.4mm}
    \renewcommand{\arraystretch}{1.12}
    \begin{tabular}{@{}l cccc@{\hspace{0.35em}} cccc@{\hspace{0.35em}} cccc@{}}
        \toprule
        & \multicolumn{4}{c}{\textbf{Qwen3-1.7B}}
        & \multicolumn{4}{c}{\textbf{Qwen3-4B}}
        & \multicolumn{4}{c}{\textbf{Qwen3-8B}} \\
        \cmidrule(lr){2-5}\cmidrule(lr){6-9}\cmidrule(lr){10-13}
        \textbf{Method}
        & \shortstack{\textbf{AIME}\\\textbf{2024}}
        & \shortstack{\textbf{AIME}\\\textbf{2025}}
        & \shortstack{\textbf{HMMT}\\\textbf{2025}}
        & \textbf{Mean}
        & \shortstack{\textbf{AIME}\\\textbf{2024}}
        & \shortstack{\textbf{AIME}\\\textbf{2025}}
        & \shortstack{\textbf{HMMT}\\\textbf{2025}}
        & \textbf{Mean}
        & \shortstack{\textbf{AIME}\\\textbf{2024}}
        & \shortstack{\textbf{AIME}\\\textbf{2025}}
        & \shortstack{\textbf{HMMT}\\\textbf{2025}}
        & \textbf{Mean} \\
        \midrule
        Base
        & 47.50 & 36.67 & 20.83 & 35.00
        & 71.67 & 67.50 & 41.67 & 60.28
        & 78.33 & 65.00 & 41.67 & 61.67 \\
        SFT$^{\dagger}$
        & 48.40 & 36.30 & 22.70 & 35.80
        & 70.20 & 62.30 & 43.40 & 58.63
        & 72.30 & 64.20 & 42.90 & 59.80 \\
        GRPO$^{\dagger}$
        & 51.10 & 38.30 & 23.70 & 37.70
        & 75.60 & 68.10 & 44.40 & 62.70
        & 76.40 & 68.90 & 46.70 & 64.00 \\
        \midrule
        OPSD
        & 54.17 & 39.17 & 27.50 & 40.28
        & 75.83 & 65.00 & 46.67 & 62.50
        & 78.33 & 64.17 & 47.50 & 63.33 \\
        EOPD
        & 56.67 & 35.83 & 22.50 & 38.33
        & 75.00 & 70.83 & 39.17 & 61.67
        & 75.83 & 66.67 & 46.67 & 63.06 \\
        DASH
        & 57.50 & 41.67 & 26.67 & 41.94
        & 75.00 & 67.50 & 43.33 & 61.94
        & 75.83 & 69.17 & 47.50 & 64.17 \\
        PW-OPSD
        & 55.00 & 41.67 & 30.00 & 42.22
        & 76.67 & 68.33 & 49.17 & 64.72
        & 80.83 & 70.83 & 44.17 & 65.28 \\
        AVSD
        & 48.33 & 32.50 & 25.83 & 35.56
        & 76.67 & 62.50 & 45.83 & 61.67
        & 80.00 & 70.00 & 45.83 & 65.28 \\
        ROSD
        & 55.83 & 40.83 & 28.33 & 41.67
        & 75.83 & 66.67 & 46.67 & 63.06
        & 78.33 & 68.33 & 46.67 & 64.44 \\
        \midrule
        \textbf{RC-OPD (Ours)}
        & 54.17 & 47.50 & 30.83 & \textbf{44.17}
        & 78.33 & 69.17 & 50.83 & \textbf{66.11}
        & 78.33 & 76.67 & 45.83 & \textbf{66.94} \\
        \bottomrule
    \end{tabular}
\end{table}
Table~\ref{tab:main-results} summarizes reasoning accuracy across three model sizes and benchmarks.
\textbf{1) Overall Performance.} Among methods evaluated with Avg@4, \texttt{RC-OPD} achieves the highest mean accuracy at all three model sizes. It improves mean accuracy by 9.66\%, 5.78\%, and 5.70\% relative to the OPSD on Qwen3-1.7B, Qwen3-4B, and Qwen3-8B, respectively, showing consistent gains across model scales.
\textbf{2) Distillation Trap.}
Baselines such as DASH and PW-OPSD vary distillation strength across tokens and achieve improvements over OPSD. These gains support adaptive allocation of supervision and are consistent with the \emph{distillation trap}: uniformly constraining all tokens may interfere with useful reasoning. Figure~\ref{fig:root-error-selectivity} further examines whether \texttt{RC-OPD} preserves valid reasoning while improving corrective alignment.
\textbf{3) Reasoning Mismatch.}
OPSD's reference-conditioned teacher may follow a path the student has not established. ROSD introduces trajectory-specific reflections and localized distillation, improving mean accuracy over OPSD at all three model sizes. \texttt{RC-OPD} further tests repairs through student continuation and provides distinct guidance for Error Stages and valid prefixes, achieving higher mean accuracy than ROSD across all sizes.
\Needspace{5\baselineskip}
\subsection{Variant Comparison}
\label{sec:ablations}

\begin{wraptable}{l}{0.50\textwidth}
    \vspace{-6pt}
    \centering
    \caption{Comparison of \texttt{RC-OPD} and its training variants. ``w/o'' denotes the omitted component. Mean averages the three benchmarks. Variant settings are described in Appendix~\ref{app:ablation-details}.}
    \label{tab:ablations}
    \vspace{3pt}
    \footnotesize
    \setlength{\tabcolsep}{2.5pt}
    \renewcommand{\arraystretch}{1.2}
    \setlength{\dashlinedash}{1.5pt}
    \setlength{\dashlinegap}{1.5pt}
    \resizebox{\linewidth}{!}{%
    \begin{tabular}{@{}llcccc@{}}
        \toprule
        \multicolumn{2}{c}{Variants} & \multicolumn{4}{c}{Avg@4 (\%)} \\
        \cmidrule(lr){1-2}\cmidrule(l){3-6}
        \# & Method & \shortstack{AIME2024} & \shortstack{AIME2025} & \shortstack{HMMT2025} & Mean \\
        \midrule
        \ding{172} & w/o Anchor supervision & 57.50 & 42.50 & 27.50 & 42.50 \\
        \hdashline
        \ding{173} & w/o Validation & 54.17 & 39.17 & 28.33 & 40.56 \\
        \hdashline
        \ding{174} & w/o Fallback supervision & 57.50 & 42.50 & 24.17 & 41.39 \\
        \hdashline
        \ding{175} & Full \texttt{RC-OPD} & 54.17 & 47.50 & 30.83 & \textbf{44.17} \\
        \bottomrule
    \end{tabular}%
    }
    \vspace{-6pt}
\end{wraptable}

We compare three training variants of \texttt{RC-OPD} on Qwen3-1.7B (Table~\ref{tab:ablations}; settings in Appendix~\ref{app:ablation-details}). \textbf{All three variants perform worse than the full method on mean accuracy}, with the largest decline observed without Iterative Counterfactual Validation. Removing valid-prefix supervision or reference fallback also lowers performance. This pattern is consistent with the complementary roles of the three components: validating repairs through student continuation, supporting valid reasoning before the Error Stage, and providing reference guidance when local repair fails. The full method combines these forms of supervision and achieves the strongest overall result.

\WFclear
\Needspace{6\baselineskip}
\subsection{Analyses}
\label{sec:further-analysis}


\begin{figure}[!htbp]
    \centering
    \includegraphics[width=\linewidth]{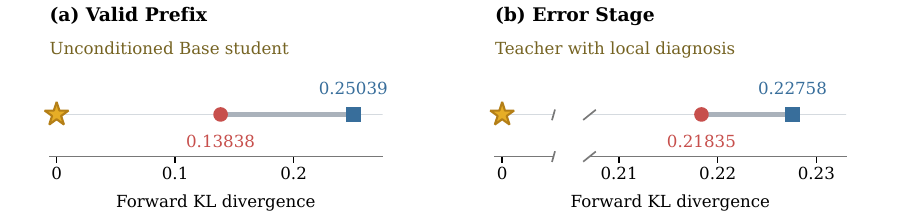}
    \caption{Valid-prefix preservation (left) and error-stage alignment (right). Blue squares denote OPSD and red circles denote \texttt{RC-OPD}; yellow stars mark the respective reference distributions at zero divergence. Both panels report forward KL from the marked reference to the evaluated model; lower is closer. The right panel uses an explicit axis break.}
    \label{fig:root-error-selectivity}
\end{figure}

\paragraph{Does RC-OPD mitigate reasoning mismatch?}
\label{sec:reasoning-mismatch-analysis}
Reference-conditioned guidance may depart from the student's reasoning path. We measure how closely \texttt{RC-OPD} preserves Base distributions on the student's valid prefixes. We compare \texttt{RC-OPD} and OPSD on the valid prefixes of the same incorrect Base trajectories from the three evaluation benchmarks. Forward KL from Base measures distributional drift; lower values indicate better preservation of valid reasoning. Detailed settings are provided in Appendix~\ref{app:selectivity-analysis}. Figure~\ref{fig:root-error-selectivity} (left) shows that \texttt{RC-OPD} reduces valid-prefix KL from Base by 44.7\% relative to OPSD, with lower divergence on all trajectories. This indicates less disruption of the student's established reasoning and provides evidence of reduced reasoning mismatch.

\paragraph{Does RC-OPD mitigate the distillation trap?}
\label{sec:distillation-trap-analysis}
The distillation trap concerns whether local supervision effectively targets erroneous reasoning. We compare \texttt{RC-OPD} and OPSD at the Error Stages of the same incorrect Base trajectories. Forward KL from a diagnosis-conditioned teacher measures alignment with corrective guidance; lower values indicate closer alignment. As shown in Figure~\ref{fig:root-error-selectivity} (right), \texttt{RC-OPD} reduces Error Stage KL by 4.1\% relative to OPSD (see Appendix~\ref{app:selectivity-analysis} for details). Together with reduced drift on valid prefixes, this suggests that \texttt{RC-OPD} directs supervision toward errors while preserving useful reasoning. As a complementary continuation analysis, we compare models from identical valid prefixes on 100 eligible questions answered incorrectly by Base, after excluding questions used to train either model (see details in Appendix~\ref{app:shared-prefix-continuation}); models receive no reference answers or diagnostic hints. \texttt{RC-OPD} achieves 54.75\% final-answer accuracy, compared with 46.50\% for OPSD and 30.25\% for Base (Table~\ref{tab:shared-prefix-continuation}; Appendix~\ref{app:shared-prefix-continuation}). The 8.25-point difference from OPSD describes this selected subset and does not establish an overall performance advantage.

\Needspace{17\baselineskip}
\paragraph{Does Iterative Counterfactual Validation improve recovery?}
\label{sec:validation-analysis}
\begin{wrapfigure}{l}{0.46\textwidth}
    \vspace{-6pt}
    \centering
    \includegraphics[width=\linewidth]{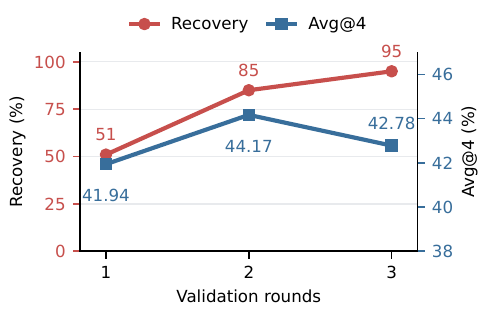}
    \caption{Effect of validation rounds on frozen-Base recovery (red, left axis) and final Avg@4 of models trained with the corresponding validation budget (blue, right axis).}
    \label{fig:multiround-repair-gain}
    \vspace{-4pt}
\end{wrapfigure}

Figure~\ref{fig:multiround-repair-gain} presents two related experiments with different settings. \textbf{1) Successive repairs improve recovery.} We use frozen Qwen3-1.7B Base to generate one response per training-pool question and retain 100 incorrect responses. Each undergoes up to three repair rounds without parameter updates or final-test data. Recovery is the fraction answered correctly within a given round budget (Appendix~\ref{app:multiround-repair-analysis}). The red curve rises from 51\% to 85\% and 95\%, showing continued recovery with diminishing gains. \textbf{2) Two validation rounds yield the best training performance.} Separately, we train \texttt{RC-OPD} with validation budgets of one, two, and three rounds. The blue curve reports the final models' Avg@4 scores of 41.94\%, 44.17\%, and 42.78\%, respectively. In \texttt{RC-OPD}, successful validation selects local supervision, so higher recovery during training means fewer trajectories receive reference-based fallback loss. The peak at two rounds suggests a balance between local repair and reference-based hindsight: retaining some fallback supervision may help students learn alternative solution paths from references.

\Needspace{14\baselineskip}
\begin{wrapfigure}{r}{0.43\textwidth}
    \vspace{-8pt}
    \centering
    \includegraphics[width=\linewidth]{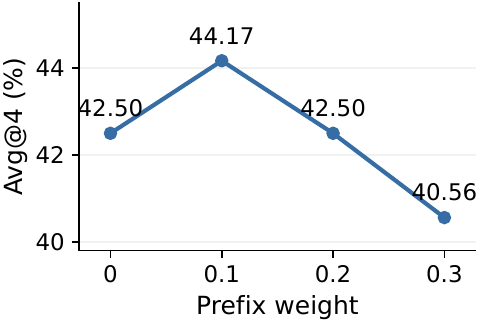}
    \caption{Sensitive Analysis on valid prefix weight $\lambda$.}
    \label{fig:prefix-weight-performance}
    \vspace{-8pt}
\end{wrapfigure}
\paragraph{How does prefix supervision weight relate to performance?}
Figure~\ref{fig:prefix-weight-performance} summarizes selected Qwen3-1.7B runs with prefix loss weights of 0, 0.1, 0.2, and 0.3. Their mean Avg@4 scores are 42.50\%, 44.17\%, 42.50\%, and 40.56\%, respectively, averaged across the three benchmarks. Weight 0.1 gives the highest score among these selected runs. Avg@4 uses four generated responses per problem. Training settings differ across entries, so this comparison is descriptive and does not isolate the effect of prefix weight.


\WFclear
\section{Conclusion}
\label{sec:conclusion}
We presented \texttt{RC-OPD}, which combines diagnosis, Iterative Counterfactual Validation, and Differentiated Distillation to address reasoning mismatch and the distillation trap. It achieves the highest mean accuracy among methods evaluated with Avg@4 across three benchmarks and three model sizes. Analyses show better corrective alignment, stronger preservation of valid prefixes, and improved recovery.

\section*{AI Use Statement}
We used generative AI tools to assist with drafting and revising the manuscript, improving clarity and organization, and preparing figures. AI assistance also supported the interpretation of experimental results. LLMs were used for error diagnosis, local repair, and response evaluation as described in the method and experimental settings. The manuscript and figures were iteratively revised based on author feedback. The authors take full responsibility for the accuracy, originality, and integrity of the paper.

\section*{Reproducibility Statement}
The method and experimental protocols are described in Sections~\ref{sec:method} and~\ref{sec:experimental-setup} and Appendix~\ref{app:experimental-details}. Additional implementation settings are provided in the supplementary configuration file.

\bibliography{iclr2027_conference}
\bibliographystyle{iclr2027_conference}

\clearpage
\appendix

\section{Additional Results and Examples}
\label{app:additional-results}
\subsection{Examples of Unresolved Reference Borrowing}
\label{app:borrowing-examples}
Figure~\ref{fig:borrowing-examples} pairs supplied reference excerpts with excerpts from the corresponding reference-conditioned responses. All three responses have correct final answers and come from problems the Base response did not answer correctly. Blue marks a reference conclusion; red marks its adoption or an explicit decision to accept it in the response. Ellipses indicate omitted text. Case identifiers are internal dataset indices, not official problem numbers. The excerpts illustrate labels assigned from the complete responses, rather than from reference mentions alone.

\begin{figure}[htbp]
\centering
\begin{minipage}{0.97\textwidth}
\textbf{(a) Accepting a count without understanding its derivation}\hfill AIME25, ID 9
\par\smallskip\hrule\smallskip
\textbf{Reference.} ``Note that we didn't consider $2$ cases but these are symmetric so the number of ways to fill the second row is \textcolor{referenceblue}{$56\cdot 6^3$}.''
\par\smallskip
\textbf{Response.} ``But why 56? [...] But without going into the detailed derivation, \textcolor{borrowred}{since the reference solution says this is the case}, and it leads to the correct prime factorization, \textcolor{borrowred}{I can accept this}.''
\par\smallskip
\emph{Missing link:} the response accepts the second-row count instead of establishing the case analysis behind~56.

\bigskip
\textbf{(b) Importing a necessary intermediate probability}\hfill AIME25, ID 12
\par\smallskip\hrule\smallskip
\textbf{Reference.} ``Thus, the probability that two segments intersect is''
\[
\frac{1}{9}+\frac{1}{18}+\frac{1}{36}+\frac{2}{9}+\frac{1}{18}
=\textcolor{referenceblue}{\frac{17}{36}}.
\]
\textbf{Response.} ``Wait, \textcolor{borrowred}{the reference solution talks about the probability that two segments intersect, which is 17/36}.'' [...] ``The probability that any two of the 25 added segments intersect is given by a detailed combinatorial and geometric analysis as \textcolor{borrowred}{$\frac{17}{36}$}.''
\par\smallskip
\emph{Missing link:} the response uses the probability in its expectation calculation without supplying the underlying combinatorial~argument.

\bigskip
\textbf{(c) Accepting the answer despite unresolved confusion}\hfill AIME24, ID 29
\par\smallskip\hrule\smallskip
\textbf{Reference.} ``Adding these up, we have $1+75+150+150+75=\textcolor{referenceblue}{451}$. Multiplying this by 2, we get $\boxed{\textcolor{referenceblue}{902}}$.''
\par\smallskip
\textbf{Response.} ``This is very confusing, but given that \textcolor{borrowred}{the reference solution arrives at 451, and the answer is 902}, \textcolor{borrowred}{I will proceed to accept this as the correct approach}.''
\par\smallskip
\emph{Missing link:} the response accepts the reference count and final answer while its interpretation of the counting argument remains~unresolved.
\end{minipage}
\caption{\textbf{Correct answers with unresolved reference dependence.} Excerpts from the supplied references and actual model responses. \textcolor{referenceblue}{Blue} identifies reference conclusions; \textcolor{borrowred}{red} identifies their adoption in the response. Mathematical and Markdown formatting is normalized for display. The missing-link descriptions summarize the full-response evidence review.}
\label{fig:borrowing-examples}
\end{figure}

\clearpage

\section{Experimental Details}
\label{app:experimental-details}

\subsection{Reference-Conditioning Analysis}
\label{app:reference-analysis}
\paragraph{Data and Evaluation.}
We evaluate frozen Qwen3-4B on AIME24 and AIME25, with 30 problems each. Base receives only the problem; the reference-conditioned variant additionally receives the worked solution and final answer through the OPSD hindsight prompt. Both generate responses from scratch.  As for
the evaluation, answer accuracy counts correct responses; filtered accuracy counts those that are both correct and free of unresolved reference borrowing. Both divide by the total number of responses. Base is evaluated without this filter.

\paragraph{Reference-borrowing assessment.}
DeepSeek-V4-Flash-0731 checks complete responses for \emph{unresolved reference borrowing}: adopting a necessary intermediate conclusion or final answer from the reference without deriving or independently verifying it. Each response is assessed twice, with disagreements reviewed.

\subsection{Variant Settings}
\label{app:ablation-details}
Table~\ref{tab:ablations} compares variants involving prefix supervision, repair validation, and reference fallback. Detailed settings are retained in \texttt{supplementary/ablation\_settings.md}.

\textbf{1)w/o Anchor supervision} removes the Anchor Prefix loss while retaining Error Stage supervision and the reference fallback. \textbf{2) w/o Validation} still diagnoses the earliest root error and proposes a route-preserving local repair, but directly constructs local distillation supervision once diagnosis and route checks pass, without using a student counterfactual continuation to select the repair. It retains both Error Stage and Anchor Prefix losses and follows Full's handling of unreliable diagnoses. Its base model, training-question order, update budget, optimizer, seed schedule, and evaluation protocol match Full. \textbf{3) w/o Fallback supervision} removes the OPSD fallback that uses the reference solution as hindsight when counterfactual validation fails.

\subsection{Error-Stage Alignment and Prefix Preservation}
\label{app:selectivity-analysis}
\paragraph{Trajectory selection.}
We start with 88 archived, answer-incorrect Base trajectories from AIME24, AIME25, and HMMT25, with one trajectory per problem. We retain 58 with a reliably identified first substantive error and a local repair that preserves the original reasoning route: 16, 20, and 22 from the respective benchmarks. The other 30 are excluded for reasons including incompatible stage formatting or unreliable diagnosis. All 58 enter the Error Stage analysis. Nine have no tokens preceding the Error Stage, leaving 49 for valid-prefix analysis. Diagnoses are reviewed twice and adjudicated by the same assistant using the problem, response, and final answer; earlier annotations were checked against worked references.

\paragraph{Distributional measurements.}
We use the Qwen3-1.7B RC-OPD and OPSD checkpoints from the main evaluation. The corrective teacher is frozen Base conditioned on the diagnosis and corrected-state goal, without the worked reference or full replacement stage. Teacher forcing supplies identical archived response prefixes to all models; thinking is enabled for the teacher and disabled for the evaluated models. We compute forward KL from the corrective teacher at Error Stage positions, excluding stage markers, and from Base at up to 128 deterministically and uniformly sampled valid-prefix positions. Both use the full vocabulary, natural logarithms, and no clipping. Scores are averaged over selected positions within each trajectory, then equally across trajectories.

\subsection{Continuation from Identical Valid Prefixes}
\label{app:shared-prefix-continuation}
\paragraph{Data and protocol.}
The candidate pool contains 6,591 questions with incorrect Qwen3-4B Base responses. After excluding questions used to train either compared model and reviewing eligibility, 190 questions remain.  For each evaluated response, the first substantive error is annotated and reviewed, and the valid prefix preceding that error is retained. Base, OPSD, and \texttt{RC-OPD} receive the same problem and the exact same prefix, without a reference answer, diagnostic hint, or inserted repair. All models are based on Qwen3-4B; the \texttt{RC-OPD} model corresponds to the 66.11\% mean result in Table~\ref{tab:main-results} and the OPSD corresponds to the 62.50\% mean result.

\paragraph{Sampling and scoring.}

Each model generates four continuations per problem with temperature 1.0, top-$p=0.95$, and a maximum of 8,192 generated tokens. These are generation samples, not independent training seeds. Final-answer accuracy is computed over all 400 continuations per model (1,200 in total), including truncated outputs. We evaluate final-answer correctness.

\begin{table}[htbp]
    \centering
    \caption{Final-answer accuracy from identical valid prefixes (Qwen3-4B, four continuations per problem).}
    \label{tab:shared-prefix-continuation}
    \small
    \setlength{\tabcolsep}{10pt}
    \renewcommand{\arraystretch}{1.12}
    \begin{tabular}{@{}lrc@{}}
        \toprule
        \textbf{Model} & \textbf{Correct / Total} & \textbf{Accuracy (\%)} \\
        \midrule
        Base & 121 / 400 & 30.25 \\
        OPSD & 186 / 400 & 46.50 \\
        \texttt{RC-OPD} & 219 / 400 & \textbf{54.75} \\
        \bottomrule
    \end{tabular}
\end{table}
On this selected subset, \texttt{RC-OPD} exceeds OPSD by 8.25 percentage points and Base by 24.50 points.

\subsection{Iterative Counterfactual Validation Settings}
\label{app:multiround-repair-analysis}
\textbf{Data and Evaluation.} Following the OPSD setup, we use the OpenThought dataset. For the main experiments, we identify 9,312 candidate training questions by retaining cases that Qwen3-1.7B answers incorrectly in a single attempt. Of these, 1,288 are used in the corresponding training run. For the repair experiment, we sample from the remaining 8,024 questions, generate one response per question using the frozen Qwen3-1.7B Base model, and retain 100 incorrect responses. To prevent formatting differences from affecting evaluation, we use DeepSeek-V4-Flash-0731 as an LLM judge to compare model responses against the reference answers.

\textbf{Repair procedure.} Each trajectory receives up to three repair rounds. At each round, diagnosis identifies the first substantive error in the editable suffix and proposes a local replacement that preserves the student's reasoning route. Diagnosis has access to the worked reference, while Base receives only the problem and the repaired prefix for continuation. Earlier repairs remain fixed, and Base generates one continuation per active trajectory per round using the same sampling settings and a fixed seed per question and round. Repair stops when the answer is correct, the round budget is exhausted, or a reliable diagnosis or compatible local repair cannot be obtained.

\textbf{Training comparison.} Separately from the frozen-Base repair experiment, we train Qwen3-1.7B \texttt{RC-OPD} with a maximum of one, two, or three validation rounds per repair chain. We evaluate each final model using the main Avg@4 protocol and report the mean across AIME24, AIME25, and HMMT25.

\subsection{Distillation Implementation}
\label{app:distillation-implementation}
\paragraph{Per-position loss.}
The implementation applies an upper cap to each vocabulary contribution of the teacher-to-student forward KL before summation. For teacher distribution $q$, student distribution $p$, and vocabulary $\mathcal V$, the loss used in Equation~\ref{eq:rc-opd-objective} is
\begin{equation}
    \ell(q,p)=\sum_{v\in\mathcal V}
    \min\!\left\{q(v)\log\frac{q(v)}{p(v)},\,\delta\right\},
    \qquad \delta=0.05.
    \label{eq:clipped-distillation}
\end{equation}
This clipped surrogate can be negative and changes gradients when the cap is active; it is distinct from the unmodified KL divergence used for distributional measurements. Raw forward KL is recorded separately during training. The cap is applied before token weighting and batch normalization.

\paragraph{Selected positions.}
We exclude control tokens and stage-heading spans from supervision. When an entire mathematical stage occupies the heading line, its content is retained. Local supervision accumulates the selected positions from all diagnosed rounds in a recovered chain; Fallback supervision uses the original rollout. An update is skipped when its total supervision weight is zero.

\subsection{Diagnosis Prompts}
\label{app:diagnosis-prompts}
The following summary groups the shared diagnosis instructions into four blocks. Placeholders denote the example contents, and ellipses omit formatting and execution details.

\begin{lstlisting}[style=diagnosisprompt]
[1. Input]
Problem: <problem>
Current trajectory: <student response>
Reference: <reference solution and answer>
Editable region: <stages or continuation that may be repaired>
...

[2. Locate and repair]
Scan the editable reasoning in order and identify a supported
mathematical error. Use the reference to check correctness while
following the student's viable route. Preserve the valid prefix
and all earlier repairs. Correct one complete logical step;
the remaining suffix will be regenerated by the student.
...

[3. Construct the three outputs]
Anchor Stage (replacement_body): the corrected step, including
the reasoning needed to reach a valid intermediate result.
Failure Reason & Goal (diagnosis, replacement_goal): explain the
error and state the intermediate result that should be established.
COT to Anchor (anchor_hindsight): a concise mathematical derivation
from the problem conditions to the Anchor Stage along the retained
route. Use only given conditions and justified reasoning.
...

[4. Return format]
Return one JSON object with status and repair fields.
ready: return the error location and the three outputs above.
unresolved: return no repair when evidence is insufficient.
route_break: return no repair when the route cannot be preserved.
...
\end{lstlisting}
For stage inputs, the error is located by its stage ID; for text inputs, it is located by an exact quote. The three mathematical outputs have the same roles in both formats.

\end{document}